\documentclass[sn-basic,iicol]{sn-jnl}

\usepackage{graphicx}%
\usepackage{multirow}%
\usepackage{amsmath,amssymb,amsfonts}%
\usepackage{amsthm}%
\usepackage{mathrsfs}%
\usepackage[title]{appendix}%
\usepackage{xcolor}%
\usepackage{color,soul}
\usepackage{textcomp}%
\usepackage{manyfoot}%
\usepackage{booktabs}%
\usepackage{algorithm}%
\usepackage{algorithmicx}%
\usepackage{algpseudocode}%
\usepackage{tabularx}
\usepackage{xcolor}
\usepackage{colortbl}

\def\@author#1{\g@addto@macro\elsauthors{\normalsize%
    \def\baselinestretch{1}%
    \upshape\authorsep#1\unskip\textsuperscript{%
      \ifx\@fnmark\@empty\else\unskip\sep\@fnmark\let\sep=,\fi
      \ifx\@corref\@empty\else\unskip\sep\@corref\let\sep=,\fi
      }%
    \def\authorsep{\unskip,\space}%
    \global\let\@fnmark\@empty
    \global\let\@corref\@empty  
    \global\let\sep\@empty}%
    \@eadauthor={#1}
}

\begin{document}

\title[ViDSOD-100: A New Dataset and A Baseline Model for RGB-D Video Salient Object Detection]{ViDSOD-100: A New Dataset and A Baseline Model for RGB-D Video Salient Object Detection}

\author{\fnm{Junhao} \sur{Lin$^{1}$}}
\author*{
\fnm{Lei} \sur{Zhu$^{1,2}$}
\thanks{*Corresponding author}
}
\email{leizhu@ust.hk}

\author
{\fnm{Jiaxing} \sur{Shen$^{4}$}}

\author{\fnm{Huazhu} \sur{Fu$^{5}$}}

\author{\fnm{Qing} \sur{Zhang$^{6}$}}

\author{\fnm{Liansheng} \sur{Wang$^{3}$}}

\vspace{-50mm}

\affil[1]{\orgname{The Hong Kong University of Science and Technology (Guangzhou)}}
\affil[2]{\orgname{The Hong Kong University of Science and Technology}}
\affil[3]{\orgdiv{School of Informatics}, \orgname{ Xiamen University}}
\affil[4]{\orgname{Department of Computing and Decision Sciences, Lingnan University}}
\affil[5]{\orgdiv{Agency for Science, Technology and Research (A*STAR)}, \orgname{the Institute of High Performance Computing (IHPC)}, \country{Singapore}}

\affil[6]{\orgdiv{School of Data and Computer Science}, \orgname{Sun Yat-Sen University}
\vspace{-5mm} 
}
\abstract{
With the rapid development of depth sensor, more and more RGB-D videos could be obtained. Identifying the foreground in RGB-D videos is a fundamental and important task. However, the existing salient object detection (SOD) works only focus on either static RGB-D images or RGB videos, ignoring the collaborating of RGB-D and video information. 
In this paper, we first collect a new annotated RGB-D video SOD (ViDSOD-100) dataset, which contains 100 videos within a total of 9,362 frames, acquired from diverse natural scenes. All the frames in each video are manually annotated to a high-quality saliency annotation.
Moreover, we propose a new baseline model, named attentive triple-fusion network (ATF-Net), for RGB-D video salient object detection. Our method aggregates the appearance information from an input RGB image, spatio-temporal information from an estimated motion map, and the geometry information from the depth map by devising three modality-specific branches and a multi-modality integration branch.
The modality-specific branches extract the representation of different inputs, while the multi-modality integration branch combines the multi-level modality-specific features by introducing the encoder feature aggregation (MEA) modules and decoder feature aggregation (MDA) modules.
The experimental findings conducted on both our newly introduced ViDSOD-100 dataset and the well-established DAVSOD dataset highlight the superior performance of the proposed ATF-Net.This performance enhancement is demonstrated both quantitatively and qualitatively, surpassing the capabilities of current state-of-the-art techniques across various domains, including RGB-D saliency detection, video saliency detection, and video object segmentation.
Our data and our code are available at 
\href{https://github.com/jhl-Det/RGBD_Video_SOD}{ViDSOD-100}.
}


\vspace{-10em}



\maketitle
\begin{figure*}[!t]
\centering
\includegraphics[width=\textwidth]{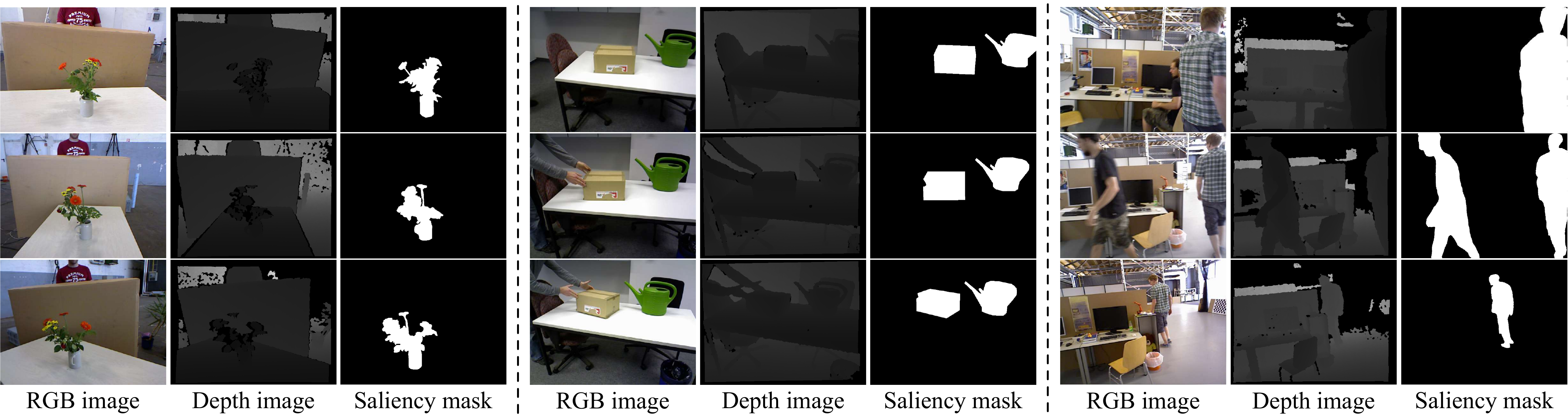}
\caption{
        The examples of proposed RGB-D video salient object detection dataset (ViDSOD-100) with pixel-level annotations.
        }
\label{fig:data_example}
\end{figure*}

\section{Introduction}
\label{sec:intro}
\begin{figure}[!t]
\centering
\includegraphics[width=\columnwidth]{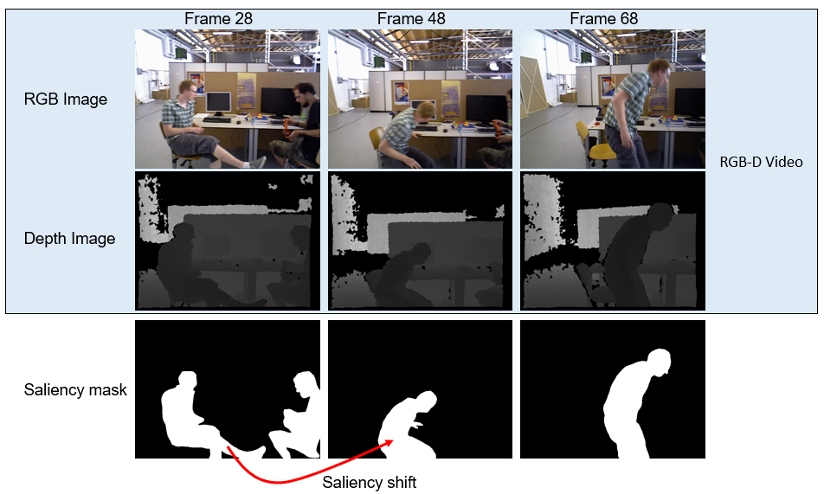}
\caption{
        Saliency shift example of our dataset (ViDSOD-100).
        }
\label{fig:saliency_shift}
\end{figure}

Salient object detection (SOD) aims to automatically detect the most visually distinguishable foreground from images. 
It benefits diverse vision tasks, including image understanding, action recognition, person re-identification, and semi-supervised/unsupervised learning.  
Due to the availability of depth sensors in modern smartphones, depth maps are capable to provide geometric and spatial information and are robust to the lighting changes, thereby enhancing SOD performance~\citep{Cong2018_Review,zhang2021rgb,zhou2021specificity}. 
In recent years, exploration of additional depth data for SOD via so-called RGB-D saliency detectors has attracted significant research attentions.
Early RGB-D salient object detection (SOD) methods often examined hand-crafted features from a pair of RGB-D images, but tended to fail in handling complex scenes, 
since the assumptions of these heuristic priors are not always correct.
Later, convolutional neural networks (CNNs)~\citep{liu2019salient,chen2018progressively, piao2019depth, fan2020rethinking, fu2020jl, zhang2020select, fan2020bbs, li2020cross,zhang2021rgb,zhou2021specificity} have been developed to learn the complementary information between the RGB image and the depth image for RGB-D SOD and achieved superior performance over traditional methods based on hand-crafted features.
However, these CNN-based methods are trained on RGB-D images and thus degrade the SOD performance in detecting salient objects of dynamic videos due to a lack of temporal information encoded in video frames.
Although many video SOD networks~\citep{wang2015saliency,chen2017video,wang2017video, Li2019MotionGA,Cong2019_TIP, ren2020tenet, zhao2021weakly, zhang2021dynamic} learned spatio-temporal information for detecting salient objects, they are trained on annotated RGB videos, which totally neglect the depth data. 
And much few works have been explored to learn a CNN for addressing the task of  RGB-D video saliency detection, which aims to identify salient objects of each video frame by leveraging paired RGB and depth video frames. 
The reason behind is a lack of an annotated dataset for RGB-D video saliency detection.
Unlike salient object detection (SOD) for static RGB-D images, salient object detection in dynamic RGB-D videos presents a greater challenge. The reason behind is that salient object(s) of different frames of the same video may dynamically vary in RGB-D videos, which is often named saliency shift~\citep{Fan_2019_CVPR_DAVSOD}. As illustrated in Fig.~\ref{fig:saliency_shift}, you can observe that from frame 28 to frame 48, the saliency in the RGB-D video transitions from two persons to just one person.
In this work, \textbf{first}, we collect a new video salient object detection (ViDSOD-100) dataset, which contains 100 videos with 9,362 video frames and 390 seconds duration, covering diverse salient object categories, and different salient object numbers.
All the video frames are carefully annotated with a high-quality pixel-level salient object mask. \textit{To the best of our knowledge, this is the first learning-oriented dataset for RGB-D video salient object detection,} which could facilitate the community to explore further in this field. 
\textbf{Second}, we develop a new baseline model, attentive triple-fusion network (ATF-Net), for this task by fusing the appearance information of the input RGB image, the temporal information of an estimated flow map, and the geometry information of the input depth image. 
In our ATF-Net, we devise three modality-specific branches to obtain multi-level CNN features learned from a depth map, a motion map, and a RGB image, and a multi-modality integration branch to aggregate three kinds of features for leveraging the temporal information and the complementary information of the depth and RGB modalities.
Then, a modality-specific encoder feature aggregation (MEA) module is devised to fuse encoder features of three images, and a modality-specific decoder feature aggregation (MDA) is designed to integrate three kinds of decoder features for detecting salient objects. 
\textbf{Finally},
we present a comprehensive evaluation of 19 state-of-the-art models on our ViDSOD-100 dataset, and experimental results show that our network clearly outperforms all the competitors, including still-image RGB-D SOD methods, video saliency detection methods and video object segmentation methods.
{\em We will release the collected dataset, our code, and our results upon the publication of this work.}

\section{Related Works}
\label{sec:related}
\vspace{2mm}
\noindent

\textbf{Image salient object detection. (SOD)}
In recent years, remarkable strides on salient object detection have been achieved by employing UNet-like feature aggregation architectures to attain outstanding performance; please refer to a recent work~\citep{Wang_2023_CVPRMENet} for a detailed review on image SOD. EGNet ~\citep{Zhao_2019_ICCVEGNet} utilizes complementary information of the salient edges and salient objects for detecting salient objects of the input image. CANet ~\citep{CANET} establishes cohesive links between each pixel and its immediate global and local contexts to identify salient objects. ICON~\citep{ICON} takes a pioneering stance to introduce a variety of feature aggregation techniques, channel enhancement for data integrity, and part-whole verification mechanisms for SOD. MENet~\citep{Wang_2023_CVPRMENet} adopted the boundary sensibility, content integrity, iterative refinement, and frequency decomposition mechanisms for SOD. Although these works have achieved superior performance for image salient object detection, they tend to fail in video RGB-D SOD due to ignore the depth information, and the temporal information of video frames.

\vspace{2mm}
\noindent
\textbf{RGB-D image salient object detection} 
endeavors to detect salient objects within intricate scenes by harnessing additional geometric information derived from depth images.
Chen et al.~(\citeyear{chen2018progressively}) presented complementarity-aware fusion (CA-Fuse) modules, while Piao et al.~(\citeyear{piao2019depth}) developed depth refinement blocks to progressively fuse RGB features and depth features. 
Fan et al.~(\citeyear{fan2020rethinking}) filtered out the noise of the depth map by a depth-depurator network for a better cross-modality feature integration. 
Zhang et al.~(\citeyear{zhang2020select}) utilized conditional variational autoencoders to approximate human annotation uncertainty and then embedded it into a probabilistic RGB-D saliency detection network. 
Li et al.~(\citeyear{li2020cross}) fused low-level, middle-level, and high-level RGB and depth features by adopting a three-level Siamese encoder-decoder structure for saliency detection in still RGB-D images. 
Zhang et al.~(\citeyear{zhang2021rgb}) learned mutual information to explicitly learn the multi-modal information between RGB data and depth data for RGB-D saliency detection. 
Zhou et al.~(\citeyear{zhou2021specificity}) designed a specificity-preserving network (SP-Net) to utilize both the shared information and modality-specific properties for RGB-D saliency detection in still images.
Recently, Lee et al.~(\citeyear{SPSN}) proposed a novel superpixel prototype sampling network that enhances the model's robustness to inconsistencies between RGB images and depth maps while also removing the influence of non-salient objects.
Cong et al. ~(\citeyear{CIR-Net}) proposed a method that facilitates cross-modality interaction and refinement through the use of attention-guided integration and refinement units in the encoder, decoder, and middleware stages.
Although working well in static images, these RGB-D saliency detectors almost neglect the temporal video information, thereby hindering the generalization capability for RGB-D video salient object detection.

\vspace{2mm}
\noindent
\textbf{Video Salient Object Detection (VSOD)} utilizs both temporal and spatial information to detect salient objects of RGB videos~\citep{Fan_2019_CVPR_DAVSOD,wang2015saliency,chen2017video,ren2020tenet,zhao2021weakly}.
Wang et al.~(\citeyear{wang2017video}) developed a pioneering fully convolutional network for detecting salient objects in videos.
Li et al.~(\citeyear{Li2019MotionGA}) presented a multi-task motion guided network to integrate still RGB image and an optical flow image for video salient object detection.
Yan et al.~(\citeyear{Yan2019SemiSupervisedVS}) embedded a non-locally enhanced recurrent module and pseudo-labels into a VSOD network (i.e., PCRNet).
Gu et al.~(\citeyear{Gu2020PyramidCS}) designed a network (i.e., PyramidCSA) grouping a set of Constrained Self-Attention (CSA) operations in Pyramid structures to capture motion information of multi-scale objects at various speeds for identifying salient objects of RGB videos.
Zhang et al.~(\citeyear{zhang2021dynamic}) addressed VSOD by developing a dynamic context-sens\-itive filtering network (DCFNet) equipped with a dynamic context-sensitive filtering module (DCFM) and a bidirectional dynamic fusion strategy.
Although achieving dominated results in VSOD, these networks mainly extracted spatio-temporal information from only RGB videos.
Since depth maps provide geometric and spatial information, the saliency detection accuracy can be enhanced by incorporating depth videos. 

\vspace{2mm}
\noindent\textbf{Video Object Segmentation (VOS)} aims to identify primary foreground objects of videos.
Oh et al.~(\citeyear{Oh2019VideoOS}) utilize a memory network (STM) to read relevant past frames stored in a memory to refine features of the current video frame for resolving VOS.
Cheng et al.~(\citeyear{cheng2022xmem}) developed a memory potentiation algorithm that systematically consolidates actively used working memory elements into long-term memory. This approach helps to prevent memory explosion and minimize performance decay in long-term prediction.
Liu et al. ~(\citeyear{QDMN_eccv2022}) conducted an evaluation of the segmentation quality of each frame in their study. This approach allowed for the selective storage of accurately segmented frames in the memory bank, which served to prevent the accumulation of errors.
Compared to these networks trained on only RGB videos, our work focuses on saliency detection in RGB-D videos, where additional depth data provides complementary information of RGB videos for saliency detection.

\section{ViDSOD-100 Dataset}
With the rapid development of Salient Object Detection (SOD), numerous datasets have been introduced. There are a plenty of SOD datasets
\citep{liu2010learning,alpert2011image,achanta2009frequency,movahedi2010design,cheng2014global,shi2015hierarchical,li2014secrets,li2015visual,wang2017learning, zhang2015salient,li2017instance,xia2017and,fan2018salient,zhang2019capsal,zeng2019towards},
which are highly representative and widely used for training or benchmarking or collected with speciﬁc properties. However, none of these datasets is for RGB-D video. 
To address this, we collect a new dataset for RGB-D video salient object detection, named ViDSOD-100. Our dataset includes 100 videos with diverse content, varying length, and pixel-level annotations. 
Some examples are shown in Figure~\ref{fig:data_example}. We will provide more details of our ViDSOD-100 from the following key aspects.

\begin{table*}[!t]
\centering
\caption{Video sources of our ViSOD-100 dataset.}
\resizebox{\linewidth}{!}{
\begin{tabular}{ccccc}
\hline
Source & \#Videos & \#Frames & Resolution & Original task \\
\hline
URFD~\citep{kwolek2014human} & 4 & 370 & 640$\times$480 & Human fall detection\\

HMP3D~\citep{lai2014unsupervised} & 14 & 1400 & 640$\times$480 & 3D scene labeling\\

BackFlow~\citep{wang2019hand} & 13 & 1300 & 640$\times$480 & 3D object scanning\\

PTB~\citep{song2013tracking} & 33 & 2791 & 640$\times$480 & Visual object tracking\\

RGB-D SLAM~\citep{sturm12iros} & 36 & 3494 & 640$\times$480 & RGB-D SLAM systems evaluation\\

\hline
\end{tabular}
}
\label{tab:videosources}
\end{table*}

\begin{table*}[!t]
\centering
\caption{Statistic Analysis of existing benchmark SOD datasets and our ViDSOD-100 dataset. 
\textbf{\#Vi.:} number of videos.
\textbf{\#Obj.:} number of salient objects in the images/video frames. 
\textbf{\#Labeled:} number of labeled images or frames.
}
\resizebox{\linewidth}{!}{
\begin{tabular}{cc|cc|ccc|ccc}
\hline
 &
 Dataset & Year & Pub. & Type & \#Vi. & \#Labeled &\#Obj. &  Resolution (H$\times$W) \\
\hline
  \multirow{8}{*}{\rotatebox{90}{Image-based}}
 & STERE~\citep{STERE} 
 & 2012 & CVPR  &  RGB-D Image & 0
 & 1,000 &  1 
 & [251,1200] $\times$ [222,900] \\

 & GIT~\citep{GIT} 
 & 2013 & BMVC & RGB-D Image & 0
 & 8 & $>1$ 
 & 640 $\times$ 480 \\

 & LFSD~\citep{LFSD} 
 & 2014 & CVPR  & RGB-D Image& 0
 & 1,00 & 1 
 & 360 $\times$ 360 \\

 & DES~\citep{DES} 
 & 2014 & ICIMCS  & RGB-D Image & 0
 & 135 & 1 
 & 640 $\times$ 480  \\
 
 & NLPR~\citep{NLPR} 
 & 2014 & ECCV  & RGB-D Image & 0
 & 1,000 & $>1$
 & 640 $\times$ 480, 480 $\times$ 640 \\

 & NJU2K~\citep{NJU2K} 
 & 2014 & ICIP  & RGB-D Image & 0
 & 1,985 & 1
 & [231,1213] $\times$ [274,828] \\
 
 & SSD~\citep{Zhu_2017_ICCV_SSD} 
 & 2017 & ICCV-W   & RGB-D Image & 0
 & 80 & $>1$
 & 960 $\times$ 1080 \\

 & SIP~\citep{fan2020rethinking} 
 & 2020 & TNNLS  & RGB-D Image & 0
 & 929 & $>1$ 
 & 992 $\times$ 744 \\

 & COME15K~\citep{come15k}
 & 2021 & ICCV  & RGB-D Image & 0
 & 15,625 & $>1$
 & [360,1280] $\times$ [360,1280]  \\

& ReDWeb-S~\citep{ReDWeb-S}
& 2022
& TPAMI
& RGB-D Image
& 0
& 3,179
& $>1$ 
& [132,772] $\times$ [153,923] \\

\hline

\multirow{6}{*}{\rotatebox{90}{Video-based}}
 & SegV2~\citep{ICCV-SegV2} 
 & 2013 & ICCV  & RGB Video & 14
 & 1,065 & $>1$ 
 & [212,360] $\times$ [259,640] \\
 
 & ViSal~\citep{TIP_ViSal} 
 & 2015 & TIP  & RGB Video& 17
 & 193 & $>1$ 
 & [240,288] $\times$ [320,512] \\

 & DAVIS~\citep{Perazzi_2016_CVPR_DAVIS} 
 & 2016 & CVPR  & RGB Video& 50 
 & 3,455  & $>1$ 
 & [900,1080] $\times$ [1600,1920]  \\

 & VOS~\citep{TIP_VOS} 
 & 2018 & TIP  & RGB Video& 200
 & 7,467 & $>1$ 
 & [321,800] $\times$ [408,800] \\
 
 & DAVSOD~\citep{Fan_2019_CVPR_DAVSOD} 
 & 2019 & CVPR & RGB Video& 226
 & 23,938 & $>1$
 & 360 $\times$ 640 \\
 
 & \textbf{ViDSOD-100 (ours)} 
 & - & - & \textbf{RGB-D Video}& \textbf{100}
 & \textbf{9,362} & \textbf{$>1$} 
 & \textbf{480 $\times$ 640}\\
 
\hline
\end{tabular}
}
\label{tab:datasets_cmp}
\end{table*}

\subsection{Data Collection}
To provide a solid basis for RGB-D video salient object, the dataset should (a) cover diverse realistic scenes and (b) contain sufficient challenging cases. The salient object would change throughout so\-me of the videos. 
As shown in Table~\ref{tab:videosources}, the videos we collected are from 5 widely-used RGB-D video datasets (i.e., URFD~\citep{kwolek2014human},  HMP3D~\citep{lai2014unsupervised}, BackFlow~\citep{wang2019hand}, RGB-D SLAM~\citep{sturm12iros}, and PTB~\citep{song2013tracking}).
These video datasets are not originally designed for salient object detection (i.e., PTB~\citep{song2013tracking} is proposed for visual object tracking), although they have the corresponding depth map in every frame. 
We then manually trim the videos to ensure each frame has at least one salient object and remove dark-screen transitions. 
Eventually, our RGB-D video dataset contains 100 video sequences, with a total of 9,362 frames. The longest video contains 100 frames, and the shortest contains 20 frames.

For each video frame, we provide a pixel-level binary saliency mask manually. 
In the realistic scenario, human observers’ attention would change according to scenario changes (see examples in Figure~\ref{fig:data_example}). 
Moreover, the number of salient objects in a video is not limited to one and would change throughout the video. 
Five human annotators are pretrained and instructed to carefully annotate all the salient objects by tracing the object boundaries. Then, two viewers are assigned to have an in-depth check to ensure that the salient object in every frame is annotated correctly.

We split the dataset into a training set and a test set according to the ratio of 6:4 and make sure challenging cases are presented in both training and test sets. This can facilitate the consistent use of and fair comparison of different methods on our dataset.

\subsection{Dataset Features and Statistics}
\begin{figure*}
\centering
\includegraphics[width=\textwidth]{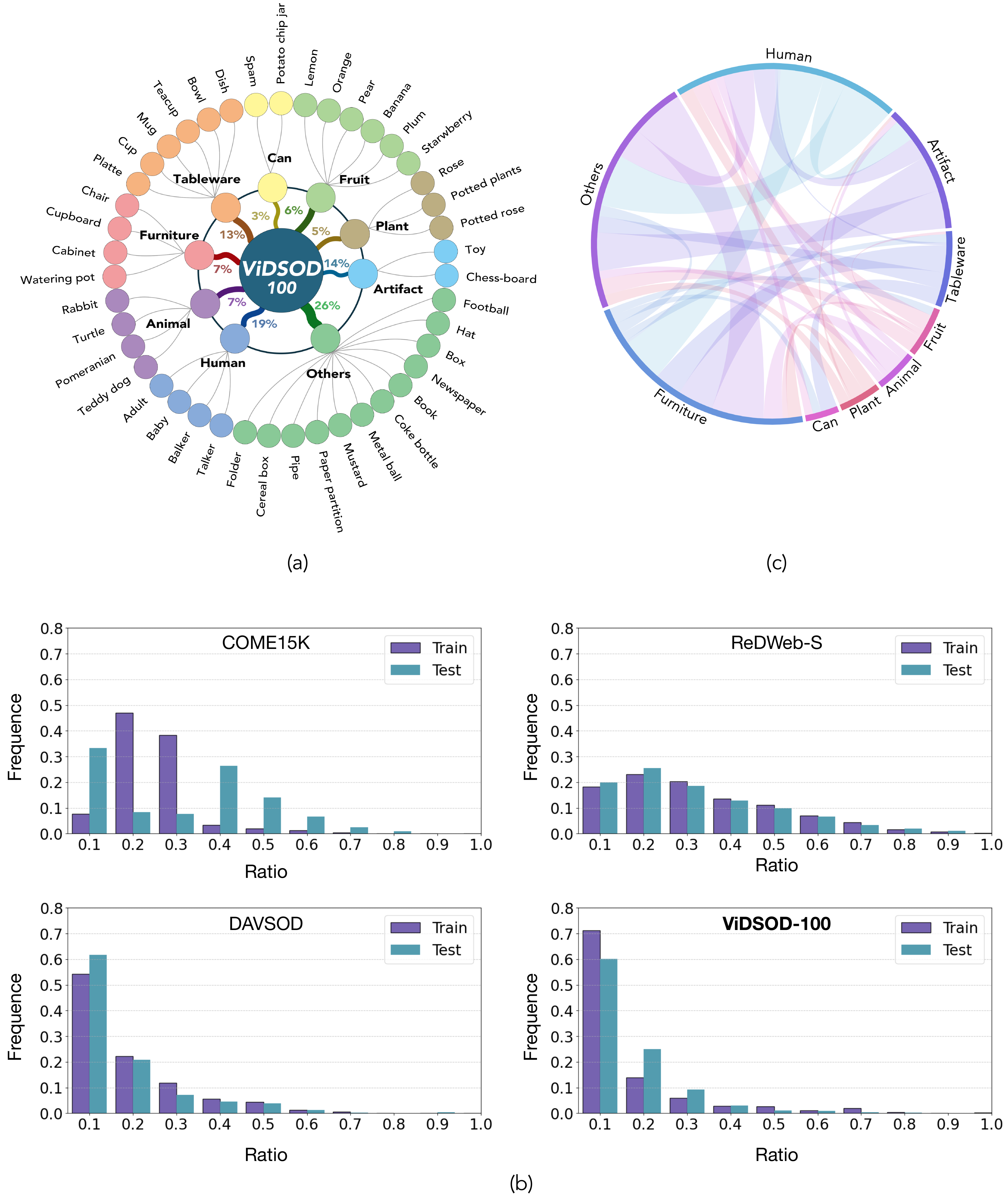}
\caption{
\textbf{Statistics of the proposed ViDSOD-100.}
(a) Salient object categories. 
(b)
Distribution of salient object ratios across our ViDSOD-100 dataset and three other expansive VOD datasets.
(c) Co-emergence of different categories in (a).
}
\label{fig:datasets_distribution}
\end{figure*}

\noindent
\textbf{Sufficient Salient Object Diversity.}
In our dataset, the salient object comprises nine main categories: Human, Animal, Furniture, Tableware, Can, Fruit, Plant, Artifact, and Others. Figure~\ref{fig:datasets_distribution}~(a) and (c) show these categories (9 main classes with 43 sub-classes) and their mutual dependencies, respectively. Besides, 30 videos have more than one salient object, making this dataset challenging for the video salient object detection task.

\vspace{3mm}
\noindent
\textbf{Dateset distribution.}
The ratio distribution of salient object sizes in frames are 0.243\% to 100\% (avg.: 11.063\%), as shown in Figure~\ref{fig:datasets_distribution}~(b), yielding a broad range.
Figure~\ref{fig:datasets_distribution}~(b) shows the distribution of other three public SOD benchmark datasets. From these figures, we can find that the salient object sizes in COME15K~\citep{come15k}, ReDWeb-S~\citep{ReDWeb-S}, and DAVSOD~\citep{Fan_2019_CVPR_DAVSOD} are predominantly concentrated within the range of 0 to 0.3. Moreover, Interestingly, our ViDSOD-100 dataset exhibited a frequency distribution, which is highly similar to that of the DAVSOD~\citep{Fan_2019_CVPR_DAVSOD} dataset.

\vspace{3mm}
\noindent
\textbf{Motion of Camera and Objects.}
As a video dataset, our ViDSOD-100 contains sufficient motion diversity for objects and cameras, mainly divided into three main types: (a) 35 videos have the salient object stay relatively static to the background the camera moves. (b) 53 videos have the salient object move, but the camera stays relatively static. (c) In the remaining 12 videos, the moving object exhibits drastic changes and/or motion blur led by camera shaking or movement. 

\begin{figure}[!t]
\centering
\includegraphics[width=\columnwidth]{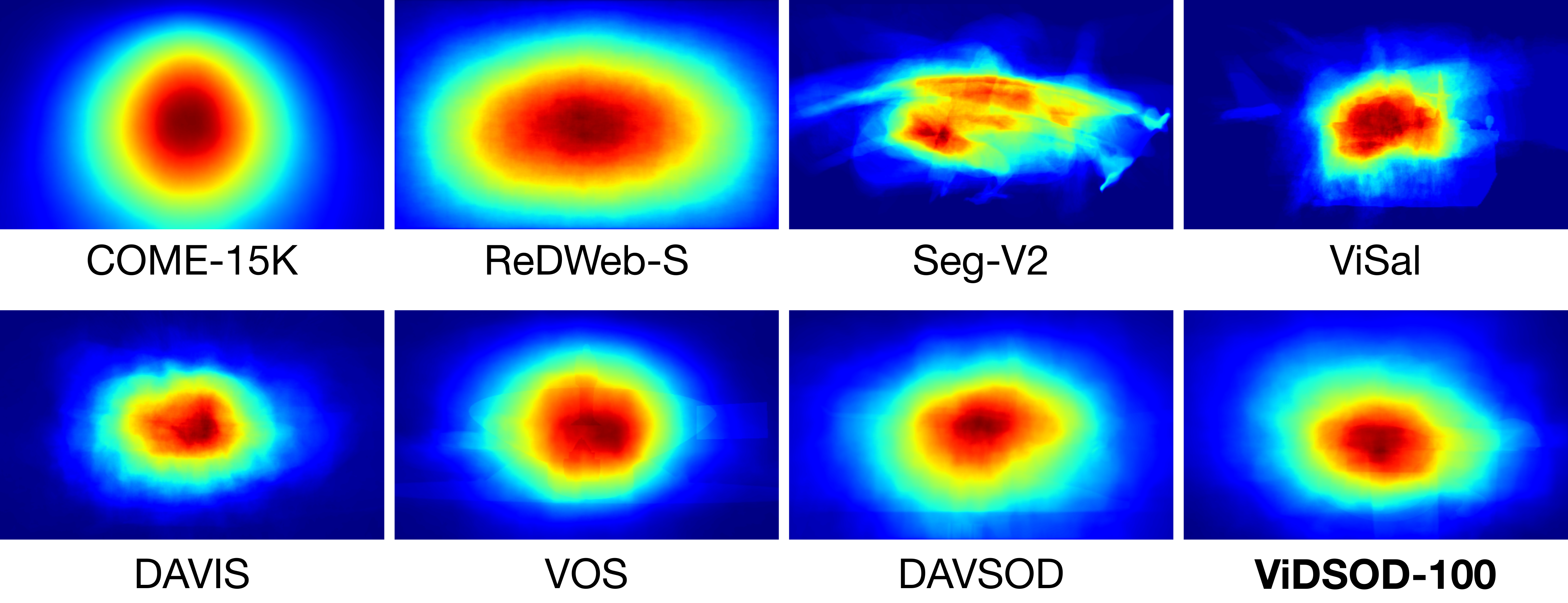}
\caption{
 Center bias of our ViDSOD-100 and existing SOD datasets.
}
\label{fig:center_bias}
\end{figure}

\vspace{3mm}
\noindent
\textbf{Center Bias.}
We visualize the center bias by computing the average saliency map across all video frames for each dataset. 
From the Fig.~\ref{fig:center_bias}, we can find that our center bias distribution closely aligns with that of DAVSOD~\citep{Fan_2019_CVPR_DAVSOD}.

\subsection{Dataset Comparisons}
Table~\ref{tab:datasets_cmp} compares the proposed ViDSOD dataset against existing datasets, including widely-used eight RGB-D image saliency object detection (SOD) datasets and five RGB video SOD datasets.

\noindent
\textbf{RGB-D Image SOD.}
There are ten benchmark datasets for RGB-D image SOD, and they are STERE~\citep{STERE}, GIT~\citep{GIT}, LFSD~\citep{LFSD}, DES~\citep{DES}, NLPR~\citep{NLPR}, NJU2K~\citep{NJU2K}, SSD~\citep{Zhu_2017_ICCV_SSD},  SIP~\citep{fan2020rethinking}, COME15K~\citep{come15k} and ReDWeb-S~\citep{ReDWeb-S}. 
Among them, STERE~\citep{STERE} was the first collection of stereoscopic photos in the field of RGB-D Image SOD. 
GIT~\citep{GIT}, LFSD~\citep{LFSD}, and DES~\citep{DES} are three small-sized datasets. 
Although the RGB-D paired images in these datasets are severely restricted by their small scales or low spatial resolution. 
To overcome these barriers, Peng et al.~\citeyear{NLPR} presented a NLPR with a large-scale RGB-D dataset with a resolution of 640 $\times$ 480, while Ju et al.~\citeyear{NJU2K} collected NJU2K, which has become one of the most popular RGB-D datasets.
SSD~\citep{Zhu_2017_ICCV_SSD} dataset partially remedied the resolution restriction of NLPR and NJU2K. However, it only contains 80 images. 
Fan et al.~\citeyear{fan2020rethinking} created a SIP dataset, which includes 929 images with object/instance-level saliency annotation.
Recently, Liu et al.\citeyear{ReDWeb-S} curated 3,179 images featuring diverse real-world scenes accompanied by high-quality depth maps. 
In parallel, Zhang et al.\citeyear{come15k} contribute a substantial RGB-D saliency detection dataset, encompassing 15,625 labeled and 5,000 unlabeled samples, catalyzing advancements in fully/weakly/unsupervised RGB-D saliency detection.
However, these datasets only explore RGB-D image pairs from static scenes for saliency detection.
Unlike them, our ViDSOD dataset in this work addresses the task of collecting and annotating saliency regions of each video frame of RGB-D videos captured from dynamic scenes.   

\vspace{3mm}
\noindent
\textbf{RGB Video SOD.}
On the other hand, there are five benchmark datasets for RGB video saliency detection. 
There are
SegV2~\citep{ICCV-SegV2}, ViSal~\citep{TIP_ViSal}, DAVIS~\citep{Perazzi_2016_CVPR_DAVIS}, VOS~\citep{TIP_VOS}, and DAVSOD~\citep{Fan_2019_CVPR_DAVSOD}.
SegV2~\citep{ICCV-SegV2} contains 14 videos of birds, animals, cars, and humans and has 1,065 video frames with dense saliency annotation.
ViSal~\citep{TIP_ViSal} is a pioneering video dataset that intends to provide a deeper exploration of video-based SOD. 
It contains 17 videos about humans, animals, motorbikes, etc. 
Each video includes 30 to 100 frames, in which salient objects are annotated according to the semantic classes of videos.
DAVIS~\citep{Perazzi_2016_CVPR_DAVIS} is a well-known video segmentation dataset about humans, animals, vehicles, objects, and actions (50 videos, with 3,455 densely saliency annotated frames). 
VOS~\citep{TIP_VOS} is the first large-scale dataset (200 videos, with 116,103 frames in total in which 7,467 frames are annotated).
It covers a wide variety of real-world scenarios and contains salient objects, which are unambiguously defined and annotated. 
DAVSOD~\citep{Fan_2019_CVPR_DAVSOD} is a recent larger-scale dataset containing 226 videos with 23,938 annotated frames. 
Beside providing object-level salient object annotations, DAVSOD also has the instance-level salient object annotations.
Although providing saliency masks in dynamic videos, these five datasets only relied on the RGB image for detecting the saliency of each video frame.
On the contrary, we collected a new dataset for RGB-D video SOD (denoted as ViDSOD-100) and annotated saliency mask of each video frame.  
Our ViDSOD-100 enables us to train CNNs for saliency detection at each video frame by utilizing RGB videos and paired depth videos.
\section{Proposed Method}
\label{sec:methods}
\begin{figure*}[!t]
\centering
\includegraphics[width=1.0\textwidth]{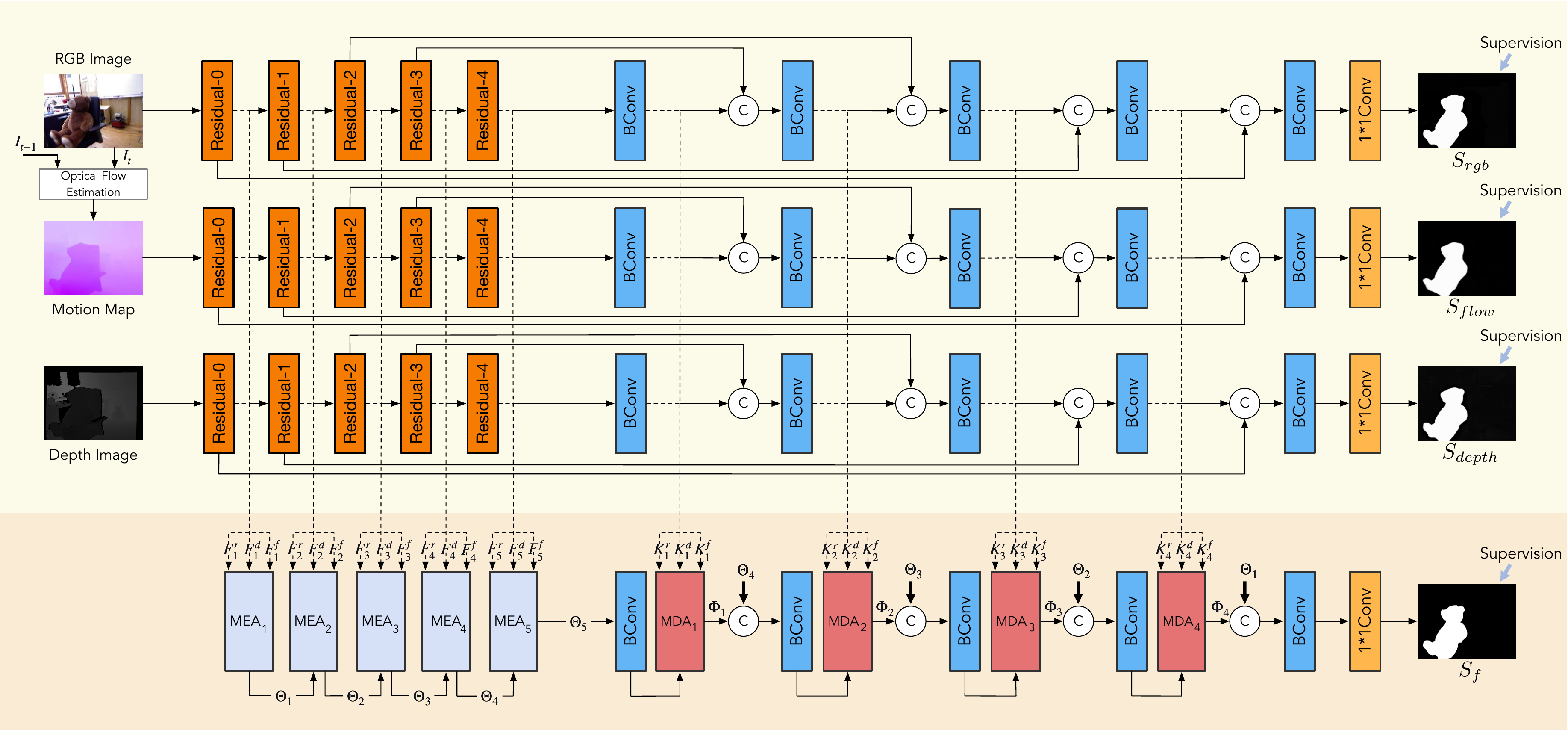}
\caption{
\textbf{The schematic illustration of our ATF-Net for RGB-D video saliency detection.}
Our ATF-Net contains three modality-specific branches and one multi-modality integration branch to fuse the appearance, temporal, and geometry information from the input RGB image, an estimated flow map, and  depth image. Moreover, five modality-specific encoder feature aggregation (MEA) modules and four modality-specific decoder feature aggregation (MDA) modules are devised to integrate multi-level features at the encoders and the decoders of three modality-specific branches. 
\textit{``C''} denotes the concatenation operation. \textit{``$BConv$''} represents a sequential operation containing a $3\times3$ convolution layer, a batch normalization, and a $ReLU$ activation function. \textit{``$1*1 \ Conv$'' is a $1\times1$ convolution layer. }
}
\label{fig:framework}
\end{figure*}

Figure~\ref{fig:framework} shows the schematic illustration of our attentive triple-fusion network (ATF-Net) for RGB-D video salient object detection.  
Our network begins by utilizing RAFT~\citep{teed2020raft} to estimate the optical flow map between the input $t$-th RGB frame and the previous ($t-1$)-th RGB frame for obtaining a motion map.
The intuition behind our network is to attentively integrate the appearance, temporal motion, and depth information from three kinds of images (RGB image, motion map, and depth image) for detecting salient objects of each video frame.
As shown in Figure~\ref{fig:framework}, our network consists of four branches: an appearance branch, a motion branch, a depth branch, and a multi-modality aggregation branch.
Taking U-Net as the basic framework, the three modality-specific branches adopt a feature extraction network with five residual blocks as the encoder to produce a set of feature maps with different spatial resolutions, utilize a decoder to fuse features at two adjacent CNN layers, and predict three saliency detection maps (see $S_{rgb}$, $S_{flow}$, $S_{depth}$ of Figure~\ref{fig:framework}) from the decoder features with the largest spatial resolutions.
Moreover, in the last multi-modality integration branch of our network, we devise a modality-specific encoder feature aggregation (MEA) module at each encoder layer to integrate encoder features from three images (i.e., a RGB image, a motion map, and a depth image), while a modality-specific decoder feature aggregation (MDA) module is formulated to aggregate three modality-specific features at each decoder layer.
Also, we predict a saliency detection map $S_f$ from the output features of the last MDA module, as shown in Figure~\ref{fig:framework}.

\subsection{Three Modality-specific Branches}
\label{sec:method-mln}

\vspace{2mm}
\noindent
\textbf{Encoder.}
The encoder at each modality-specific branch utilizes a feature extraction backbone (i.e., Res2Net-50~\citep{gao2019res2net} pre-trained on ImageNet \citep{russakovsky2015imagenet}) with five residual blocks. 
We employ $\{F_i^{r} \mid  i \in [1, 5]\}$ to denote five feature maps from the RGB image, $\{F_i^{f}\mid 
 i\in [1,5]\}$ to denote five feature maps from the flow map, and $\{F_i^{d}\mid i\in [1,5]\}$ to stand for five feature maps from the input depth map.

\noindent
\textbf{Decoder.}
After obtaining five encoder features for each input image, we then follow the U-Net structure to progressively fuse features at two adjacent layers.
Let $F_5$, $F_4$, $F_3$, $F_2$, and $F_1$ denote five encoder feature maps of a modality-specific branch, while the five decoder feature maps are denoted as $D_5$, $D_4$, $D_3$, $D_2$, and $D_1$. 
To compute these five decoder feature maps, we first pass the feature map $F_5$ at deepest encoder CNN layer to a ``BConv'' block to estimate $D_5$. 
The ``BConv'' block consists of a $3$$\times$$3$ convolutional layer, a batch normalization layer, and a ReLU activation layer.
Then, we up-sample $D_5$, concatenate the up-sampled result with $F_4$, and then pass the concatenation result into another ``BConv'' block to compute $D_4$.
Similarly, we can progressively compute $D_3$, $D_2$, and $D_1$, and then predict a saliency map from the feature map $D_1$.
And these predicted saliency map at three branches are denoted as $S_{rgb}$, $S_{flow}$, and $S_{depth}$, respectively.


\subsection{Multi-modality Integration Branch}
\label{sec:method-integration}

In contrast to these static-image RGB-D saliency detection methods, we incorporate a motion map to learn a spatio-temporal feature representation for boosting saliency detection in videos.
Furthermore, the geometric information in the depth data enables the network to work well for scenarios when the foreground salient objects share similar appearance to the non-salient backgrounds, and the robustness of additional depth data for the light change also benefits saliency detection.
Motivated by this, we develop a deep model to incorporate the RGB image, the motion map, and the depth image for addressing the task of RGB-D video salient object detection.
To do so, we devise a modality-specific encoder feature aggregation (MEA) module to fuse encoder features learned from the RGB image, the motion map, and the depth image, while modality-specific decoder feature aggregation (MDA) module is designed to aggregate decoder features learned from three inputs.

\begin{figure}[!t]
\centering
\includegraphics[width=1\columnwidth]{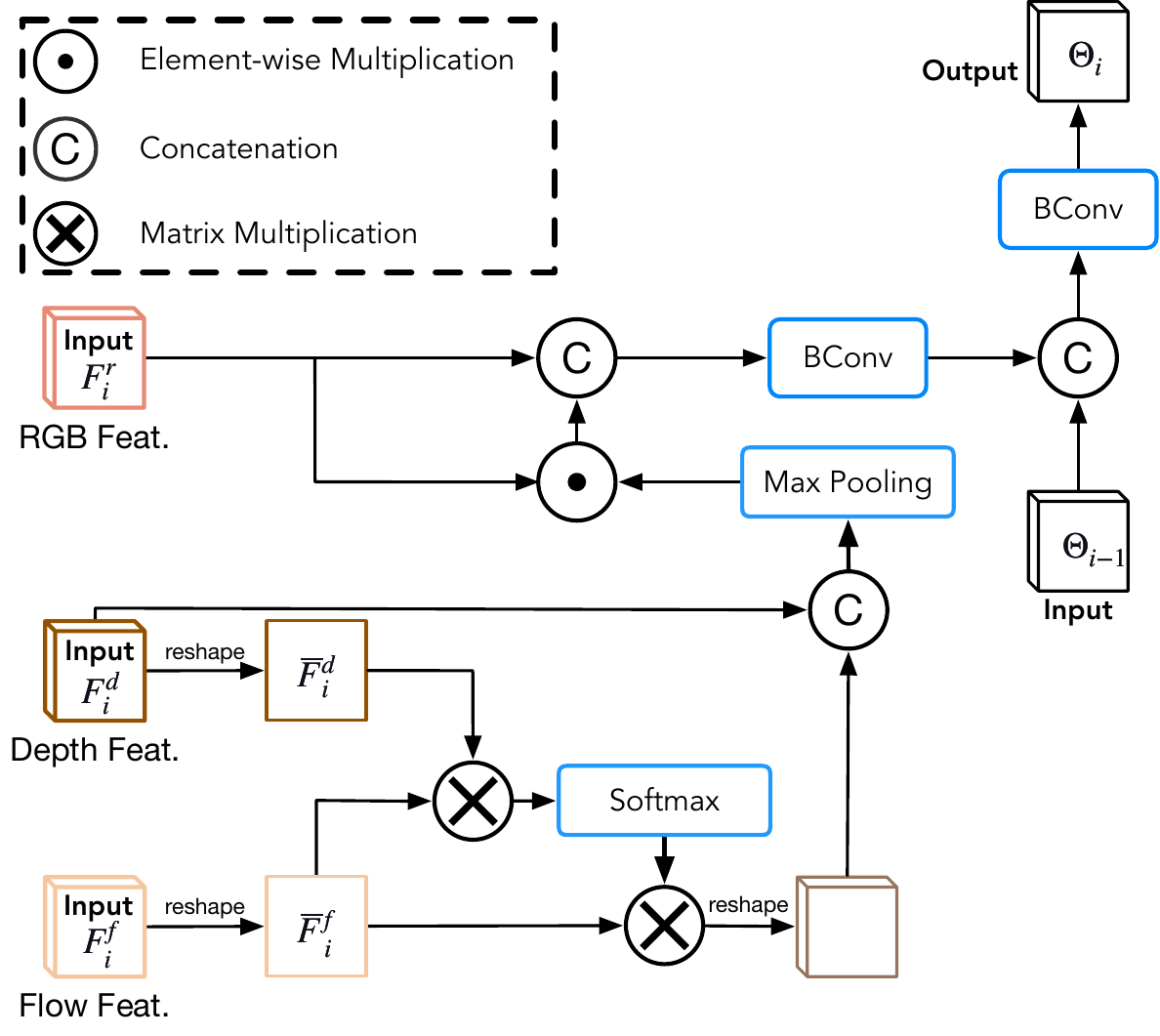}

\caption{
The schematic illustration of the $i$-th modality-specific encoder feature aggregation (MEA) module.
It takes the encoder RGB feature $F_i^{r}$, the encoder motion feature $F_i^{f}$, and the encoder depth feature $F_i^{d}$ as three inputs and the output feature map is $\Theta_{i}$. }
\label{fig:mfi}
\end{figure}

\subsubsection{Modality-specific Encoder Feature Aggregation (MEA) Module}
As shown in Figure~\ref{fig:framework}, we design five modality-specific encoder feature aggregation (MEA) modules (see $\textit{MEA}_1$, $\textit{MEA}_2$, $\textit{MEA}_3$, $\textit{MEA}_4$, and $\textit{MEA}_5$) to integrate three kinds of features (i.e., RGB encoder features, the motion encoder features, and the depth encoder features) at each encoder layer. 
Figure~\ref{fig:mfi} shows the schematic illustration of $i$-th MEA module  ($1 \leq i \leq 5$).
It takes three features at the $i$-th encoder features from three modality specific branches and the output feature $\Theta_{i-1}$ of the ($i-1$)-th MEA module as the input, and outputs an aggregated feature map $\Theta_{i}$.
The reason of introducing $\Theta_{i-1}$ is to propagate the aggregated features of ($i-1$)-th MEA module into the next MEA module. 
Note that $\textit{MEA}_1$ does not have any $\Theta_{i-1}$; see Figure~\ref{fig:framework}.

Specifically, we first let $ F_i^{r}$  $\in$ $\mathbb{R} ^{c \times w \times h} $,
$ F_i^{f}$  $\in$ $\mathbb{R} ^{c \times w \times h} $, and
$ F_i^{d}$ $\in$ $\mathbb{R} ^{c \times w \times h} $ denote the input RGB feature, the input motion feature, and the input depth feature, respectively.
Then, we utilize a $1$ $\times$ $1$ convolution layer on $F_i^{r}$ to reduce its feature channel number to be a half, and the new feature is $\overline{F}_i^{r}$ $\in$ $\mathbb{R} ^{\frac{c}{2} \times w \times h}$.
Then, regarding $F_i^{f}$ and $F_i^{d}$, we apply a $1$ $\times$ $1$ convolution layer on $F_i^{f}$ to reduce a half of the feature channel number and reshape it to obtain two  feature maps, i.e., $\overline{F}_i^{f}$ $\in$ $\mathbb{R} ^{\frac{c}{2} \times (wh)}$ and $\overline{F}_i^{d}$ $\in$ $\mathbb{R} ^{(wh) \times \frac{c}{2}}$.
After that, to fuse the encoder depth feature map and the encoder flow feature map, we multiply $\overline{F}_i^{d}$ and $\overline{F}_i^{f}$ to compute a similarity map, apply a Softmax function on the similarity map to obtain a feature map $\mathcal{A}$, multiple $\mathcal{A}$ with $\overline{F}_i^{f}$, concatenate the multiplication result with $\overline{F}_i^{d}$, and apply a max pooling operation on the concatenated feature map to produce a fused feature map $\mathcal{B}$, as:
\begin{equation}
\label{eq:phi}
    \begin{split}
    \mathcal{B} &= MP(Concat ( F_i^{d} \ , 
    reshape(\overline{F}_i^{f} \times \mathcal{A} ) ) ) \ , \\
    \mathcal{A} &= Softmax (  \overline{F}_i^{d} \times \overline{F}_i^{f} ) \ , 
    \end{split}
\end{equation}
where $MP$ denotes a maximum pooling layer with kernel size $3\times3$, 
and $Concat$ is to concatenate two feature maps along the feature channel dimension.
$reshape$ is to reshape a $\frac{c}{2}$$\times$$wh$ 2D matrix into a $\frac{c}{2}$$\times$$w$$\times$$h$ 3D feature map.
$Softmax$ represents a softmax operation, and $\times$ is a matrix multiplication. 
Then, we incorporate the RGB color feature map by multiplying the obtained $\mathcal{B}$ from Eqn.~\eqref{eq:phi} with $\overline{F}_i^{r}$, and then concatenated the multiplication result with $\overline{F}_i^{r}$, and pass the concatenation result into a ``BConv'' block to produce a feature map $\mathcal{C}$.
Finally, to further integrate the output feature map $\Theta_{i-1}$ of ($i-1$)-th MEA module, we concatenate $\mathcal{C}$ with $\Theta_{i-1}$ and we pass the concatenated result into a ``BConv'' block to produce the final result $\Theta_{i}$ of $i$-th MEA module.
Hence, we compute $\Theta_{i}$ as follows:
\begin{equation}
\label{Eq:Theta}
    \begin{split}
    \Theta_{i}  
    &= Bconv( Concat(\Theta_{i-1}, \mathcal{C}) )  \ , \\
    \mathcal{C} 
    &= Bconv ( Concat(\overline{F}_i^{r} \odot \mathcal{B} , \overline{F}_i^{r}) ) \ ,
    \end{split}
\end{equation}
where $\odot$ denotes the element-wise matrix multiplication.

\begin{figure*}[!t]
\centering
\includegraphics[width=1.0\textwidth]{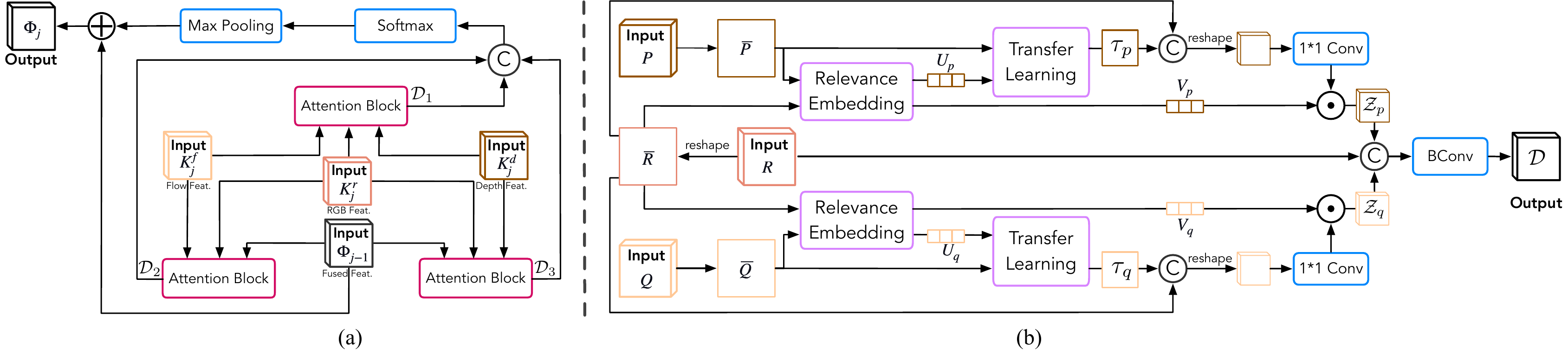}
\caption{
The schematic illustration of j-th our modality-specific decoder feature aggregation (MDA) module in (a) and its attention block in (b). The MDA module in (a) takes the decoder RGB feature $K_j^{r}$, the decoder motion feature $K_j^{f}$, the decoder depth feature $K_j^{d}$, and the output feature $\Phi_{j-1}$ of the (j-1)-th MDA module as the four inputs, and outputs the feature $\Phi_{j}$. The attention block in (b) takes $P$, $R$, and $Q$ as the inputs and outputs $\mathcal{D}$. Here $R$ is $K_j^{r}$ of (a). $P$ and $Q$ are two features from the feature set \{$K_j^{f}$, $K_j^{d}$, and $\Phi_{j-1}$\} of (a).
}
\label{fig:acf}
\end{figure*}
\subsubsection{Modality-specific Decoder Feature Aggregation (MDA) Module}

Moreover, we devise a modality-specific decoder feature aggregation (MDA) module to integrate three decoder features (i.e., RGB decoder features, the motion decoder features, and the depth decoder features) at each decoder layer, and our network thus contains four MDA modules (i.e., $\textit{MDA}_1$,  $\textit{MDA}_2$, $\textit{MDA}_3$, and $\textit{MDA}_4$ of Figure~\ref{fig:framework}). 
Figure~\ref{fig:acf}(a) shows the schematic illustration of $j$-th MDA module ($1 \leq j \leq 4$), which takes the decoder feature $K_j^{r}$ from the RGB image, the decoder feature $K_j^{f}$ from the motion map, the decoder feature $K_j^{d}$ from the depth map, and the output feature map $\Phi_{j-1}$ of the ($j-1$)-th MDA module as the input, and outputs an aggregated feature map $\Phi_{j}$.
In $\textit{MDA}_1$, we compute $\Phi_{1}$ by passing the output feature map $\Theta_{5}$ of $\textit{MEA}_5$ into a ``BConv'' block; see Figure~\ref{fig:framework}.

The intuition behind our MDA module is to transfer these three decoder features $K_j^{f}$, $K_j^{d}$, and $\Phi_{j-1}$ to enhance the RGB decoder feature $K_j^{r}$ for saliency detection. 
To do so, our MDA module adopts three attention blocks to refine $K_j^{r}$ by selecting two feature maps from the feature set\{ $K_j^{f}$, $K_j^{d}$, $\Phi_{j-1}$\}.
The first attention block utilizes $K_j^{f}$ and $K_j^{d}$, while the second attention block considers $K_j^{f}$ and $\Phi_{j-1}$. 
Moreover, $K_j^{d}$ and $\Phi_{j-1}$ are selected into the third attention block to refine the RGB decoder feature map $K_j^{r}$.
Here, we utilize $\mathcal{D}_1$, $\mathcal{D}_2$, and $\mathcal{D}_3$ to denote output features of three attention blocks.
After that, we concatenate $\mathcal{D}_1$, $\mathcal{D}_2$, and $\mathcal{D}_3$, pass the concatenated result into a softmax function and a max-pooling operation, and add the pooling result with the output feature $\Phi_{j-1}$ of our $(j-1)$-th MDA module to compute the output feature map $\Phi_{j}$ of our $j$-th MDA module:
\begin{equation}
\label{Eq:MDA}
    \Phi_{j} = \Phi_{j-1} + \textit{MP}(\textit{Softmax}( \textit{Concat}(\mathcal{D}_1, \mathcal{D}_2, \mathcal{D}_3))) \ , 
\end{equation}
where $MP$ denotes a max-pooling operation with kernel size $3\times3$.

\subsubsection{Attention Blocks} 
Figure~\ref{fig:acf}(b) shows the schematic illustration of the attention block of our MDA block (see Figure~\ref{fig:acf}(a)).
The attention block transfers two input features to refine the RGB decoder feature map $K_j^{r}$ (denoted as $R$ $\in$ $\mathbb{R} ^{c_1 \times w_1 \times h_1}$ for simplicity), and these two features (denoted as $P$ $\in$ $\mathbb{R} ^{c_1 \times w_1 \times h_1}$ and $Q$ $\in$ $\mathbb{R} ^{c_1 \times w_1 \times h_1}$) are randomly selected the the feature set \{ $K_j^{f}$, $K_j^{d}$, $\Phi_{j-1}$\}.
Our attention block consists of three main steps.
\textbf{First}, it reshapes the three input 3D features $R$, $P$, and $Q$ into three 2D matrices $\overline{R}$ $\in$ $c_1$$\times$$w_1h_1$, $\overline{P}$ $\in$ $c_1$$\times$$w_1h_1$, and $\overline{Q}$ $\in$ $c_1$$\times$$w_1h_1$.
Then, we pass $\overline{R}$ and $\overline{P}$ to a relevance embedding block~\citep{feng2021task} for computing a similarity map $\mathcal{S}_{p}$ (size: $w_1$$h_1$ $\times$ $w_1$$h_1$) between $P$ and $R$, and $\overline{R}$ and $\overline{Q}$ is also fed to a relevance embedding block for computing a $w_1$$h_1$ $\times$ $w_1$$h_1$ similarity map $\mathcal{S}_{q}$ between $Q$ and $R$.
\textbf{Second}, we compute a transfer attention map $\mathcal{U}_{p}$ (a vector with $w_1h_1$ elements) and a soft attention map $\mathcal{V}_{P}$ (a vector with $w_1h_1$ elements) from the similarity matrix $\mathcal{S}_{p}$:
$\mathcal{U}_{p}(m) = {argmax}_n(\mathcal{S}_{p}(m, n))$, and 
$\mathcal{V}_{p}(m) = {max}_n(\mathcal{S}_{p}(m, n))$.
Apparently, $\mathcal{U}_{p}(m)$ computes the column index for the maximum value of all elements in $m$-th row of the 2D similarity map $\mathcal{S}_{p}$, while $\mathcal{V}_{p}(m)$ denote the maximum value of all elements in $m$-th row of the 2D similarity map $\mathcal{S}_{p}$.  
Then, we generate a transferred feature map $\mathcal{T}_{p}$ ($c_1$$\times$$w_1h_1$) based on the $\mathcal{U}_{p}$ and $\overline{P}$:
$\mathcal{T}_{p}(x, y) = \overline{P}(x, \mathcal{U}_{p}(y))$.
Similarly, we compute a transfer attention map $\mathcal{U}_{q}$ and a soft attention map $\mathcal{V}_{q}$ from the similarity map $\mathcal{S}_{q}$, and then obtain a transferred feature map $\mathcal{T}_{q}$ from $\mathcal{U}_{q}$ and $\overline{Q}$.
\textbf{Lastly}, we compute the output feature map $\mathcal{D}$ of our attention block:
\begin{equation}
\label{Eq:attention_block}
\begin{split}
\mathcal{D} &= BConv(Concat(R, \mathcal{Z}_{p}, \mathcal{Z}_{q})) \ , \\
\mathcal{Z}_{p} &= Conv(reshape(Concat(\overline{R}, \mathcal{T}_{p} ))) \odot \mathcal{V}_{p} \ , \\
\mathcal{Z}_{q} &= Conv(reshape(Concat(\overline{R}, \mathcal{T}_{q} ))) \odot \mathcal{V}_{q} \ , 
\end{split}
\end{equation}
where $\mathcal{D}$, $\mathcal{Z}_{p}$, and $\mathcal{Z}_{q}$
are three matrices with the size of $c_1$$\times$$w_1$$\times$$h_1$.
$Conv$ denotes a $1$$\times$$1$ convolutional layer.
$reshape$ is to reshape a $c_1$$\times$$w_1h_1$ 2D matrix into a $c_1$$\times$$w_1$$\times$$h_1$ 3D feature map.

\subsection{Loss Function}
The total loss $L_{total}$ of our network is computed by adding the prediction error of the saliency maps (i.e., $S_{rgb}$, $S_{flow}$, $S_{depth}$, and $S_f$ of Figure~\ref{fig:framework}) at four branches:
\begin{equation}
\begin{split}
    \label{Eq:total_loss}
    L_{total} = & \Omega(S_{rgb}, GT) + \lambda_1 \Omega(S_{depth}, GT) 
    + \\ & \lambda_2 \Omega(S_{flow}, GT) + \lambda_3  \Omega(S_{f}, GT) \ ,
\end{split}
\end{equation}
where $GT$ denotes the ground truth of the saliency detection for the input RGB-D video frame.
$\Omega(S_{rgb}, GT)$,
$\Omega($$S_{flow}$, $GT)$,
$\Omega($$S_{depth}$, $GT)$, 
and
$\Omega($$S_{f}$, $GT)$ denote the loss functions of the predicted saliency maps $S_{rgb}$, $S_{flow}$, $S_{depth}$, and $S_{f}$ at four branches of our network. 
Here, we empirically use the pixel position aware loss~\citep{wei2020f3net} (a combination of a weighted binary cross entropy (wBCE) loss and a weighted IoU (wIoU) loss) to compute the loss function $\Omega$.
$\lambda_1$, $\lambda_2$, and $\lambda_3$ are the weights, which we empirically set as $\lambda_1 = \lambda_2 = \lambda_3=1$.
In the testing stage, we empirically take  $S_f$ at the multi-modality integration branch as the final result of our network.

\section{Experimental Results}
\noindent
\textbf{Evaluation metrics.}
We adopt four widely-used metrics to quantitatively compare our RGB-D video salient object detection (SOD) network against state-of-the-art methods, and they are the Mean Absolute Error (MAE)~\citep{perazzi2012saliency},
F-measure (F$_{\beta}$)~\citep{achanta2009frequency}, 
S-measure (S$_{\alpha}$)~\citep{fan2017structure}, 
and E-measure (E$_{\phi}$)~\citep{QDMN_eccv2022}. 
In general, a better RGB-D video SOD method shall have larger F$_{\beta}$, S$_{\alpha}$, E$_{\phi}$ scores, and a smaller MAE score. 
\begin{table*}[!t]
\centering
\caption{Quantitative comparisons between our ATF-Net and state-of-the-art methods for RGB-D video saliency detection.
$\ast$ denotes the use of depth maps as auxiliary inputs. 
For DAVSOD~\citep{Fan_2019_CVPR_DAVSOD} dataset, we utilize a monocular depth estimator ~\citep{rajpal2023highmde} to generate pseudo depth maps for each frame of the video.
The input images all have dimensions of $ 352 \times 352$.
}
\resizebox{\linewidth}{!}{
\begin{tabular}{l|cccc|cccc|ccc|cccc}
\hline
\multirow{2}{*}{Method} & 
\multirow{2}{*}{Year} & 
\multirow{2}{*}{Type} & 
\multirow{2}{*}{Backbone} 
& \multirow{2}{*}{Depth}
& \multicolumn{4}{c|}{ViDSOD-100} 
&
\multicolumn{3}{c|}{DAVSOD$^{\ast}$}  
&
\multirow{2}{*}{$\#$ Param.(M)} & \multirow{2}{*}{FLOPs(G)} & \multirow{2}{*}{MACs(G)} & \multirow{2}{*}{FPS}
\\
&&&&
& MAE $\downarrow$ & F$_{\beta}$ $\uparrow$ & S$_{\alpha}$ $\uparrow$ & E$_{\phi}$ $\uparrow$ &
MAE $\downarrow$ & F$_{\beta}$ $\uparrow$ & S$_{\alpha}$ $\uparrow$ &
&&& \\
\hline
UCNet\citep{Zhang2020UCNetUI}        & 2020  & image & ResNet-50 
& $\checkmark$
& 0.052 & 0.723 & 0.765 & 0.827 
& - & - & -
& 26 & 16 & 8 & 30\\
S2MA\citep{Liu2020LearningSS}        & 2020 & image & VGG-16
& $\checkmark$
& 0.056 & 0.710 & 0.768 & 0.812
& - & - & -
& 87 & 255 & 128 & 15 \\
BBS-Net\citep{Zhai2021BifurcatedBS}  & 2020  & image & ResNet-50
& $\checkmark$
& 0.036 & 0.831 & 0.829 & 0.871
& - & - & -
& 50 & 31 & 16 & 18\\
D3Net\citep{fan2020rethinking}       & 2021 & image & VGG-16
& $\checkmark$
& 0.047 & 0.771 & 0.793 & 0.841
& - & - & -
& 45 & 135 & 68 & 26 \\
DCF\citep{Ji2021CalibratedRS}        
& 2021  & image & ResNet-50
& $\checkmark$
& 0.039 & 0.805 & 0.787 & 0.827
& - & - & -
& 108 & 54 & 27 & 8 \\
TriTransNet\citep{TriTransNet}     
& 2021  & image &  ResNet-50
& $\checkmark$
& 0.039 & 0.809 & 0.790 & 0.826 
& - & - & -
& 139 & 553 & 277 & 4 \\
CMINet\citep{come15k}     
& 2021  & image & ResNet-50
& $\checkmark$
& 0.037 & 0.808 & 0.823 & 0.864
& - & - & -
& 214 & 188 & 94 & 3 \\
VST\citep{VST}     
& 2021  & image & T2T-ViT
& $\checkmark$
& 0.037 & 0.810 & 0.824 & 0.863
& - & - & -
& 84 & 200 & 100 & 10 \\
SP-Net\citep{zhou2021specificity}     
& 2021  & image & Res2Net-50
& $\checkmark$
& 0.036 & 0.813 & 0.829 & 0.874
& - & - & -
& 150 & 68 & 34 & 6 \\
CIRNet\citep{CIR-Net}       
& 2022  & image & ResNet-50
& $\checkmark$
& 0.035 & 0.812 & 0.828 & 0.872
& - & - & -
& 107 & 320 & 160 & 4 \\
SPSN\citep{SPSN}     
& 2022  & image & VGG-16
& $\checkmark$
& 0.038 & 0.809 & 0.791 & 0.829
& - & - & -
& 100 & 37 & 19 & 12 \\
\hline
STM\citep{Oh2019VideoOS}               
& 2019  & video & ResNet-50
& $\times$
& 0.032 & 0.793 & 0.846 & 0.861
& 0.076 & 0.655 & 0.735
& 39 & 43 & 22 & 15 \\
STCN\citep{STCN}              
& 2021   & video & ResNet-50
& $\times$
& 0.041 & 0.810 & 0.789 & 0.841
& 0.086 & 0.652 & 0.740
& 54 & 53 & 27 & 23 \\
QDMN\citep{QDMN_eccv2022}
& 2022   & video & ResNet-50
& $\times$
& 0.042 & 0.785 & 0.810 & 0.840
& 0.084 & 0.657 & 0.748
& 105 & 74 & 37 & 13\\
XMem\citep{cheng2022xmem}            
& 2022   & video  & ResNet-50
& $\times$
& 0.040 & 0.791 & 0.816 & 0.848
& 0.081 & 0.660 & 0.748
& 62 & 57 & 29 & 25 \\
TBD\citep{TBD-ECCV-2022}            
& 2022   & video & ResNet-50
& $\times$
& 0.040 & 0.789 & 0.816 & 0.847
& 0.082 & 0.662 & 0.749
& 9 & 41 & 21 & 28\\
\hline
STM$^\ast$\citep{Oh2019VideoOS}             
& 2019  & video & ResNet-50
& $\checkmark$
& 0.032 & 0.795 & 0.847 & 0.861
& 0.077 & 0.651 & 0.734
& 39 & 43 & 22 & 15 \\
STCN$^\ast$\citep{STCN}              
& 2021   & video & ResNet-50
& $\checkmark$
& 0.040 & 0.812 & 0.791 & 0.843
& 0.085 & 0.654 & 0.742
& 54 & 53 & 27 & 23 \\
QDMN$^\ast$\citep{QDMN_eccv2022}
& 2022   & video & ResNet-50
& $\checkmark$
& 0.040 & 0.789 & 0.820 & 0.842
& 0.080 & 0.658 & 0.749
& 105 & 74 & 37 & 13\\
XMem$^\ast$\citep{cheng2022xmem}            
& 2022   & video  & ResNet-50
& $\checkmark$
& 0.039 & 0.792 & 0.817 & 0.850
& 0.079 & 0.662 & 0.751
& 62 & 57 & 29 & 25 \\
TBD$^\ast$\citep{TBD-ECCV-2022}            
& 2022   & video & ResNet-50
& $\checkmark$
& 0.038 & 0.791 & 0.818 & 0.849
& 0.077 & 0.663 & 0.752
& 9 & 41 & 21 & 28\\
\hline
MGA\citep{Li2019MotionGA}           
& 2019  & video & ResNet-50
& $\times$
& 0.092 & 0.596 & 0.656	& 0.731
& 0.097	& 0.623 & 0.739
& 92 & 116 & 58 & 13 \\
RCRNet\citep{Yan2019SemiSupervisedVS} 
& 2019   & video & ResNet-50
& $\times$
& 0.044 & 0.750 & 0.796 & 0.865
& 0.087 & 0.653 & 0.741
& 54 & 273 & 117 & 20 \\

PyramidCSA\citep{Gu2020PyramidCS}    
& 2020   & video & MobileNetV3
& $\times$
& 0.065 & 0.754 & 0.730 & 0.758
& 0.086 & 0.655 & 0.741
& 3 & \textbf{15} & \textbf{8} & 34 \\

DCFNet\citep{zhang2021dynamic}   
& 2021   & video & ResNet-101
& $\times$
& 0.031 & 0.816 & 0.833 & 0.872
& 0.074 & 0.660 & 0.741

& 72 & 233 & 117 & 5 \\
UFO\citep{su2022unified}              
& 2022   & video & VGG-16
& $\times$
& 0.034 & 0.812 & 0.828  & 0.870
& 0.076 & 0.657 & 0.739
& 56 & 781 & 390 & 6 \\
\hline
MGA$^\ast$\citep{Li2019MotionGA}      
& 2019  & video & ResNet-50
& $\checkmark$
& 0.072	& 0.609 & 0.667 & 0.733
& 0.089	& 0.628 & 0.741
& 92 & 116 & 58 & 13 \\

RCRNet$^\ast$\citep{Yan2019SemiSupervisedVS} 
& 2019   & video & ResNet-50
& $\checkmark$
& 0.040 & 0.756 & 0.801 & 0.869
& 0.087 & 0.655 & 0.742
& 54 & 273 & 117 & 20 \\

PyramidCSA$^\ast$\citep{Gu2020PyramidCS}    
& 2020   & video & MobileNetV3
& $\checkmark$
& 0.054 & 0.761 & 0.732 & 0.768
& 0.085 & 0.658 & 0.744
& 3 & \textbf{15} & \textbf{8} & 34 \\

DCFNet$^\ast$\citep{zhang2021dynamic} 
& 2021   & video & ResNet-101
& $\checkmark$
& 0.031 & 0.824 & 0.842 & 0.883
& 0.073 & 0.662 & 0.744
& 72 & 233 & 117 & 5 \\

UFO$^\ast$\citep{su2022unified}
& 2022   & video & VGG-16
& $\checkmark$
& 0.033 & 0.816 & 0.830 & 0.876
& 0.072 & 0.661 & 0.741
& 56 & 781 & 390 & 6 \\

\hline
\textbf{ATF-Net (Ours)} 
& -  & video & MobileNetV3
& $\checkmark$
& 0.045 & 0.787 & 0.809 & 0.851
& 0.087 & 0.654 & 0.741 
& 60 & 30 & 15 & 33 \\
\textbf{ATF-Net (Ours)} 
& -  & video & VGG-16
& $\checkmark$
& 0.036 & 0.813 & 0.824 & 0.875
& 0.081 & 0.655 & 0.743 

& 97 & 212 & 106 & 16 \\
\textbf{ATF-Net (Ours)} 
& -  & video & ResNet-50
& $\checkmark$
& 0.030 & 0.834 & 0.846 & 0.885
& 0.069 & 0.667 & 0.746

& 129 & 126 & 63 & 12 \\
\textbf{ATF-Net (Ours)} 
& -  & video & Res2Net-50
& $\checkmark$
& \textbf{0.028} 
& \textbf{0.864} 
& \textbf{0.877} 
& \textbf{0.899}
& \textbf{0.061} 
& \textbf{0.673} 
& \textbf{0.752} 
& 130 & 129 & 65 & 10\\
\hline
\end{tabular}
} 
\label{tab:sota}
\end{table*}

\noindent
\textbf{Implementation Details.}
Our model is implemented in PyTorch, and trained on two GeForce RTX 2080 Ti GPUs.
The backbones (e.g., ResNet-50~\citep{he2016deep}, and Res2Net-50~\citep{gao2019res2net}) we used are pre-trained on ImageNet~\citep{deng2009imagenet}.
Since RGB, optical flow and depth images have different channels, the input channel of the depth encoder is modified to 1. We adopt the Adam algorithm to optimize the proposed model. The initial learning rate is set to $1\times 10 ^ {-4}$ and is divided by 10 every 20 epochs. The input resolutions of RGB, optical flow and depth images are resized to $352 \times 352$. 
The training images are augmented using various strategies, including rotating, random flipping, and random pepper. The batch size is set to 4 and the model is trained over 50 epochs. During the testing stage, the RGB, optical flow and depth images are resized to $352 \times 352$ and then fed into the model to obtain prediction maps. Then, bilinear interpolation is applied to upsample the prediction maps to the original size to achieve the final evaluation. Finally, the output of the multi-modality aggregation branch is the final prediction map for our model.

\vspace{2mm}
\noindent
\textbf{Comparative methods.}
To evaluate the effectiveness of our RGB-D video salient object detection network, we compare it against 18 state-of-the-art methods, including
BBS-Net~\citep{Zhai2021BifurcatedBS},
S2MA~\citep{Liu2020LearningSS},
UCNet~\citep{Zhang2020UCNetUI},
D3Net~\citep{fan2020rethinking},
DCF~\citep{Ji2021CalibratedRS},
SP-Net~\citep{zhou2021specificity},
CIRNet~\citep{CIR-Net},
SPSN~\citep{SPSN},
STM~\citep{Oh2019VideoOS},
XMem~\citep{cheng2022xmem},
STCN~\citep{STCN},
QDMN~\citep{QDMN_eccv2022},
TBD~\citep{TBD-ECCV-2022},
MGA~\citep{Li2019MotionGA},
RCRNet~\citep{Yan2019SemiSupervisedVS},
PyramidCSA~\citep{Gu2020PyramidCS},
DCFNet~\citep{zhang2021dynamic} and
UFO~\citep{su2022unified}.
Among them, BBS-Net, S2MA, UCNet, D3Net, DCF, SP-Net, CIRNet, and SPSN are developed for single-image RGB-D salient object detection, while MGA, RCRNet, PyramidCSA, DCFNet, and UFO are utilized for video RGB salient object detection (SOD).
Moreover, STM, QDMN, XMem, STCN and TBD are designed for video object segmentation.
Note that all video RGB SOD and video object segmentation methods only take the RGB images as the input.
To adapt these methods for addressing the RGB-D video saliency detection, we also modify these methods by taking the RGB image and the corresponding depth image as the input for fair comparisons.
To provide fair comparisons, we use their public implementations of all compared methods and re-train these methods on our ViDSOD-100 dataset to obtain their best performance for comparisons.
Note that their backbone network has also been pre-trained on ImageNet~\citep{deng2009imagenet}.
Moreover, we randomly select 60 videos from our ViDSOD dataset for training and the remaining 40 videos are for testing different methods.

\begin{figure*}[!t]
\centering
\includegraphics[width=\textwidth]{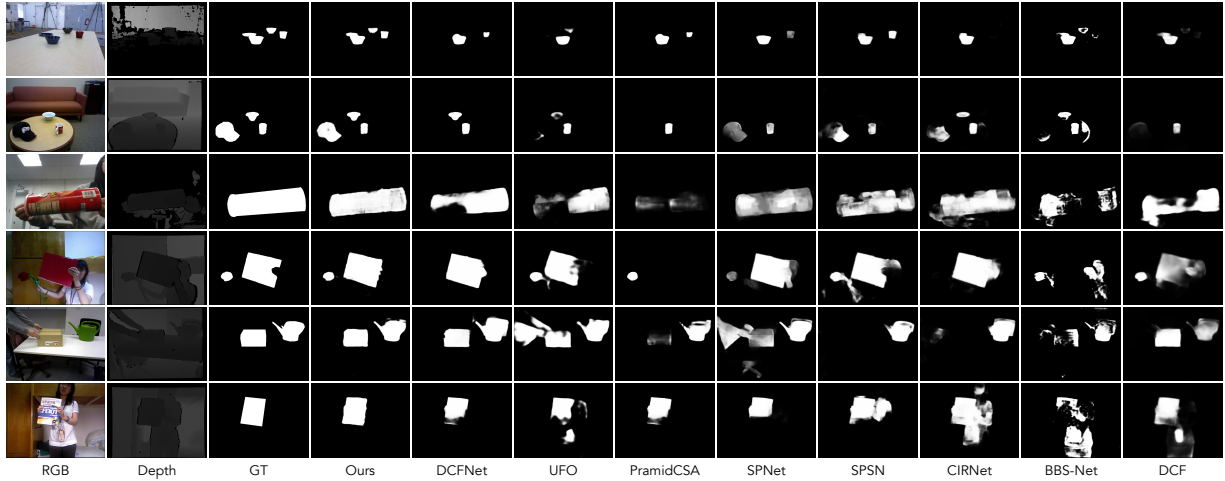}
\caption{
Visual comparisons of RGB-D video salient object detection maps produced by our network (4-th column; denotes as ``Ours'') and state-of-the-art methods (5-th to 13-th columns) against ground truths (3-rd column). Apparently, our method has a higher accuracy of detecting salient objects from RGB-D videos than all competitors. ``RGB'' and ``depth'' denote the input RGB and depth video frame. ``GT'' represents the ground truth of the saliency detection on input RGB-D video frame.
}
\label{fig:visualization}
\end{figure*}

\subsection{Comparison Against State-of-the-art Methods}
\noindent
\textbf{Quantitative comparisons.}
From the Table~\ref{tab:sota},
we have the following observations: (a) When our network utilizes different feature extraction backbones (e.g., MobileNetV3, VGG-16, ResNet-50), the FLOPs, MACs, and inference time of our network are not the same. Although our network does not the best performance in terms of the FLOPs, the MACs, and the inference time, our network outperforms all compared methods in terms of the MAE score, the F$_\beta$ score, the S$_\alpha$ score, and the E$_\phi$ score, which are 0.028, 0.864, 0.877, and 0.899. 
Specifically, i) Our ATF-Net with MobileNetV3 has larger FLOPs and MACs scores than the compared SOTA method (i.e., PyramidCSA~\citep{Gu2020PyramidCS}), while our inference time (33 FPS) is comparable to that of PyramidCSA (34 FPS). 
ii) Among all five methods taking VGG-16 as the feature extraction backbone, our method takes the 3rd rank on the FLOPs and MACs, and the 2nd rank for the inference time, and our FLOPs, MACs, and the inference time are 212 G, 106G, and 16 FPS. On contrary, our MAE, F$_\beta$ , S$_\alpha$ and E$_\phi$ scores are largest, and they are 0.036, 0.813, 0.824, and 0.875.  
iii) Regarding all methods with a backbone of ResNet-50, our FLOPs MACs, FPS take the 10-th rank among all methods with a backbone of ResNet-50, but our network can decrease the MAE score from 0.032 to 0.030; increase the F$_\beta$ score from 0.813 to 0.834; and increase the E$_\phi$ score from 0.872 to 0.885. 
iv) Regarding the backbone of Res2Net-50, our FLOPs MACs, FPS take 2nd rank, but our network improve the (MAE, F$_\beta$, S$_\alpha$, E$_\phi$) score from (0.036, 0.813, 0.829, 0.874) to (0.028, 0.864, 0.877, 0.899).

\vspace{2mm}
\noindent
\textbf{Qualitative comparisons.}
Figure~\ref{fig:visualization} visually compares RGB-D video salient object detection results produced by our network and state-of-the-arts methods. 
From the visual results, we can find that our ATF-Net (see 4-th column of Figure~\ref{fig:visualization}) can more accurately identify salient objects from RGB-D videos than all the competitors.
However, other methods tend to wrongly identify non-salient backgrounds or miss some salient objects in their results, especially for cases with multiple salient objects at the first two rows of Figure~\ref{fig:visualization}.

\begin{table*}[!t]
\centering
\caption{
Quantitative comparisons of our method and four constructed baseline networks.
\textit{``MEA''} represents the modality-specific encoder feature integration module, while \textit{``MDA''} stands for modality-specific decoder feature integration module. \textit{``attBLK''} stands for the attention block of our MDA module.
} 
\resizebox{\linewidth}{!}{
\begin{tabular}{l|ccc|ccc|ccc|ccc}
\hline
  \multirow{2}{*}{Network}
& \multirow{2}{*}{MEA}
& \multirow{2}{*}{MDA}
& \multirow{2}{*}{attBLK} 
& \multicolumn{3}{c|}{ViDSOD-100}
& \multicolumn{3}{c|}{DAVSOD$^{\ast}$}
& \multirow{2}{*}{$\#$ Param.(M) }
& \multirow{2}{*}{FLOPs(G) }  
& \multirow{2}{*}{FPS} \\
& & & 
& MAE $\downarrow$ 
& F$_{\beta}$ $\uparrow$
& S$_{\alpha}$ $\uparrow$  
& MAE $\downarrow$
& F$_{\beta}$ $\uparrow$
& S$_{\alpha}$ $\uparrow$  
& & & \\
\hline
basic & $\times$ & $\times$ &$\times$     
& 0.045 & 0.787 & 0.809
& 0.089 & 0.635 & 0.708
& \textbf{99} & \textbf{46} & \textbf{15} \\

basic+MEA &$\checkmark$ &$\times$ &$\times$  
& 0.037 & 0.841 & 0.837
& 0.081 & 0.649 & 0.736
& 130 & 57 & 12 \\
basic+MDA &$\times$ &$\checkmark$  &$\checkmark$ 
& 0.035 & 0.825 & 0.831
& 0.078 & 0.653 & 0.739
& 72 & 107 & 12 \\

\hline
Ours-w/o-attBLK &$\checkmark$ &$\checkmark$ &$\times$
& 0.032	& 0.838	& 0.839
& 0.073	& 0.662 & 0.743
& 129 & 57 & 11 \\
\hline
\textbf{ATF-Net (Ours)} &$\checkmark$ &$\checkmark$ &$\checkmark$  
& \textbf{0.028} & \textbf{0.864} & \textbf{0.877}
& \textbf{0.061} & \textbf{0.673} & \textbf{0.752}
& 130 & 129 & 10\\
\hline
\end{tabular}
}
\label{tab:ablations}
\end{table*}

\begin{table*}
\centering
\caption{
Significance of ground-truth depth maps.
} 
\begin{tabular}{l|ccccc}
\hline
  Network
& Type of depth maps
& MAE $\downarrow$ 
& F$_{\beta}$ $\uparrow$
& S$_{\alpha}$ $\uparrow$ 
& E$_{\phi}$ $\uparrow$\\
\hline
\multirow{3}{*}{ATF-Net}
& DPT ~\citep{ranftl2021visionDPT}
& 0.031 & 0.855 & 0.859 & 0.883 \\

& DPT-B+R+AL ~\citep{rajpal2023highmde}
& 0.029 & 0.860 & 0.871 & 0.894\\

& GT depth maps 
& \textbf{0.028} 
& \textbf{0.864} 
& \textbf{0.877} 
& \textbf{0.899} \\

\hline

\end{tabular}
\label{tab:effect_GT_depth}
\end{table*}

\subsection{Ablation Studies}
We extend our analysis through a series of ablation study experiments to underscore the efficacy of various inputs,
including optical flow maps and depth maps, as well as the contributions of MEA, MDA, the attention block within our network, and the integration of depth maps.
These experiments are carried out on both our proposed ViDSOD-100 dataset and the challenging video salient object detection dataset, DAVSOD~\citep{Fan_2019_CVPR_DAVSOD}.
Importantly, we employ a monocular depth estimator~\citep{rajpal2023highmde} to generate pseudo-depth images for individual video frames, and for the acquisition of optical flow maps, we utilize the RAFT~\citep{teed2020raft}.
Here, we consider four baseline networks. 
The first baseline network (denoted as ``basic'') is constructed by replacing all MEA and MDA modules from our ATF-Net with a simple concatenation operation for fusing features.  
The second baseline (denoted as ``basic+MEA'') is to add all MEA modules into ``basic'', while the third one (denoted as ``basic+MDA'') is to add all MDA modules into ``basic''.
The last baseline (denoted as ``Ours-w/o-attBLK'') is to remove only all attention blocks of our MDA modules from our network.
Table~\ref{tab:ablations} reports the quantitative results of four metrics of our method and four constructed baseline networks.

\vspace{2mm}
\noindent
\textbf{Effectiveness of MEA.}
As shown in Table~\ref{tab:ablations}, ``basic+MEA'' has larger F$_\beta$, S$_\alpha$, and E$_\phi$ scores and a smaller MAE score than ``basic'', showing that MEA modules enable our network to enhance RGB-D video saliency detection accuracy.

\vspace{2mm}
\noindent
\textbf{Effectiveness of MDA.}
From the quantitative results of Table~\ref{tab:ablations}, we can observe that ``basic+MDA'' has a superior performance on all four metrics than ``basic''.
It indicates that our MDA modules have their contributions to the success of our network for RGB-D video saliency detection.

\vspace{2mm}
\noindent
\textbf{Effectiveness of the attention block in our MDA.}
Also, we can observe that our network consistently outperforms ``Ours-w/o-attBLK'' in terms of all four metrics, which shows that the attention blocks in our MDA can well leverage the features from the flow map and the depth map, as well as the aggregated features from the previous decoder layer to refine the RGB decoder features, thereby improving the RGB-D video saliency detection performance of our ATF-Net.  

\begin{table*}[!t]
\centering
\caption{
Ablation studies on multiple inputs.
The term "simple concat" refers to the fusion of depth/optical flow images with RGB images through direct concatenation of their channel dimensions, followed by a $1 \times 1$ convolutional layer.
} 
\resizebox{\linewidth}{!}{
\begin{tabular}{l|cc|c|ccc|ccc}
\hline
  \multirow{2}{*}{Network}
& \multirow{2}{*}{Depth}
& \multirow{2}{*}{Optical flow}
& \multirow{1}{*}{Fusion} 
& \multicolumn{3}{c}{ViDSOD-100}
& \multicolumn{3}{|c}{DAVSOD$^{\ast}$} \\
& & & manner
& MAE $\downarrow$ 
& F$_{\beta}$ $\uparrow$
& S$_{\alpha}$ $\uparrow$  
& MAE $\downarrow$
& F$_{\beta}$ $\uparrow$
& S$_{\alpha}$ $\uparrow$  
\\
\hline
vanilla & $\times$ & $\times$ &$\times$     
& 0.066 & 0.751 & 0.728
& 0.167 & 0.334 & 0.553
\\

vanilla+D-concat
& $\checkmark$ 
& $\times$ 
& simple concat   
& 0.061 & 0.754 & 0.732
& 0.160 & 0.341 & 0.559
\\

vanilla+D-branch
& $\checkmark$ 
& $\times$ 
& UNet-like branch    
& 0.054 & 0.765 & 0.747
& 0.129  & 0.520 & 0.656
\\
vanilla+D-ED
& $\checkmark$ 
& $\times$ 
& MEA+MDA    
& 0.044 & 0.779 & 0.767
& 0.090 & 0.551 & 0.677
\\

vanilla+O-concat
& $\times$ 
& $\checkmark$ 
& simple concat     
& 0.062 & 0.751 & 0.730
& 0.162 & 0.339 & 0.558
\\

vanilla+O-branch
& $\times$ 
& $\checkmark$ 
& UNet-like branch 
& 0.055 & 0.761 & 0.743
& 0.131  & 0.512 & 0.646
\\

vanilla+O-ED
& $\times$ 
& $\checkmark$ 
& MEA+MDA
& 0.043 & 0.776 & 0.766
& 0.110 & 0.549 & 0.672
\\

vanilla+D-concat+O-concat 
& $\checkmark$ 
& $\checkmark$ 
& simple concat    
& 0.060 & 0.756 & 0.735
& 0.160 & 0.343 & 0.561  \\

vanilla+D-branch+O-branch 
& $\checkmark$ 
& $\checkmark$ 
& UNet-like branch    
& 0.045 & 0.787 & 0.809
& 0.089 & 0.635 & 0.708 \\

\textbf{ATF-Net (Ours)} 
& $\checkmark$ 
& $\checkmark$ 
& MEA+MDA
&\textbf{0.028} & \textbf{0.864} & \textbf{0.877}
& \textbf{0.061} & \textbf{0.673} & \textbf{0.752}
\\

\hline
\end{tabular}
}
\label{tab:ablation_multi-inputs}
\end{table*}


\begin{table*}
\centering
\caption{
Cost paid for the performance improvements.
} 
\resizebox{\linewidth}{!}{
\begin{tabular}{l|cc|ccc|ccc|ccc}
\hline
  \multirow{2}{*}{Network}
& \multirow{2}{*}{Depth}
& \multirow{2}{*}{Optical flow}
& \multicolumn{3}{c}{ViDSOD-100}
& \multicolumn{3}{|c|}{DAVSOD$^{\ast}$}
& \multirow{2}{*}{$\#$ Param.(M) }
& \multirow{1}{*}{Training}  
& \multirow{2}{*}{FPS} \\
& & 
& MAE $\downarrow$ 
& F$_{\beta}$ $\uparrow$
& S$_{\alpha}$ $\uparrow$  
& MAE $\downarrow$
& F$_{\beta}$ $\uparrow$
& S$_{\alpha}$ $\uparrow$  
& &  time cost & 
\\
\hline
vanilla & $\times$ & $\times$ 
& 0.066 & 0.751 & 0.728
& 0.167 & 0.334 & 0.553
& \textbf{24} & \textbf{6} hours & \textbf{38}
\\

vanilla+D-ED
& $\checkmark$ 
& $\times$ 
& 0.044 & 0.779 & 0.767
& 0.090 & 0.551 & 0.677
& 103 & 9 hours & 12
\\

vanilla+O-ED
& $\times$ 
& $\checkmark$ 
& 0.043 & 0.776 & 0.766
& 0.110 & 0.549 & 0.672
& 103 & 9 hours & 12 
\\

\textbf{ATF-Net (Ours)} 
& $\checkmark$ 
& $\checkmark$ 
&\textbf{0.028} & \textbf{0.864} & \textbf{0.877}
& \textbf{0.061} & \textbf{0.673} & \textbf{0.752}
& 130 & 11 hours & 10 
\\

\hline
\end{tabular}
}
\label{tab:ablation_eff_dep_motion}
\end{table*}




\subsection{More Analysis}
\textbf{Effectiveness of the depth map.}
Table~\ref{tab:ablation_multi-inputs} presents our experimental results on the importance of depth maps in ATF-Net. We evaluated the performance of ATF-Net with the depth branch removed (denoted as ``Ours-w/o-depth") and compared it against five video RGB SOD methods and five video object segmentation methods that utilize depth maps as additional input. These methods include MGA~\citep{Li2019MotionGA}, RCRNet~\citep{Yan2019SemiSupervisedVS}, PyramidCSA~\citep{Gu2020PyramidCS}, DCFNet~\citep{zhang2021dynamic}, 
UFO~\citep{su2022unified}, STM~\citep{Oh2019VideoOS}, XMem~\citep{cheng2022xmem}, STCN~\citep{STCN}, QDMN~\citep{QDMN_eccv2022}, and TBD~\citep{TBD-ECCV-2022}.
Note that these ten methods are not designed to address video RGB-D SOD, so they do not have the depth map as additional input. Therefore, we simply concatenate the depth map with RGB video frame along with the channel dimension to enable them to extract information from the depth map.
After introducing the foreground information in the depth map, we boost the performance of these methods further, which proves the depth map is helpful.

\vspace{2mm}
\noindent
\textbf{Effectiveness of the estimated depth map.}
To explore the effect of the estimated depth for the model, we construct a trivial video RGB-D dataset (DAVSOD$^{\ast}$) by adding the estimated depth map from~\citep{rajpal2023highmde} with the original video RGB dataset (DAVSOD). Table~\ref{tab:sota} shows the MAE,  F$_\beta$,  and S$_\alpha$ results of five state-of-the-art methods (MGA~\citep{Li2019MotionGA}, RCRNet~\citep{Yan2019SemiSupervisedVS},
PyramidCSA~\citep{Gu2020PyramidCS}, 
DCFNet~\citep{zhang2021dynamic} and
UFO~\citep{su2022unified}) and our ATF-Net, we can find that all five compared methods and our network suffer from a degraded performance when we change the training set from ViDSOD-100 (with the depth map from hardware devices) to DAVSOD$^{\ast}$ (with the estimated depth map). For example, our MAE is increased from 0.028 to 0.061; our F$_\beta$ score is decreased from 0.864 to 0.673; our S$_\alpha$ score is decreased from 0.877 to 0.752. And the MAE, F$_\beta$, and S$_\alpha$ scores of UFO$^{\ast}$ are degraded from (0.033, 0.816, 0.830) on ViDSOD-100 to (0.072, 0.661, 0.741) on DAVSOD$^{\ast}$. Moreover, similar to ViDSOD-100, our network also outperforms all five compared methods in terms of MAE,  F$_\beta$,  and S$_\alpha$ on DAVSOD$^{\ast}$.
Moreover, we conduct experiments on our ViDSOD-100 dataset with different qualities of the depth maps, which are the ground truth, and estimated depths from two monocular depth estimation methods. They are DPT~\citep{ranftl2021visionDPT} and its advanced iteration, DPT-B+R+AL~\citep{rajpal2023highmde}. Table~\ref{tab:sota} reports the video RGB-D SOD results of our network with different depth maps. It shows that the incorporation of higher-quality depth images undeniably led to a notable enhancement in ATF-Net's performance. However, it's crucial to note that despite these improvements, the performance still falls short when compared to using ground-truth depth maps. This underscores the undeniable significance of genuine depth information in augmenting the overall effectiveness of ATF-Net, as well as other similar methodologies.

\vspace{2mm}
\noindent
\textbf{Effectiveness of the depth and motion map.}
In Table~\ref{tab:ablation_eff_dep_motion},
we have conducted ablation studies to analyze the effectiveness of the depth map and the temporal information of the input motion map. Specifically, we construct three baseline networks by progressively removing the depth map or the motion map from our network. The first baseline (denoted as ``vanilla") is to remove the input depth map and the motion map from our network for video RGB-D SOD. The second baseline (denoted as ``vanilla+D-ED")  adds the input depth map into ``vanilla", and then we modify our MEA and MDA blocks to fuse features from the input RGB image and the input depth image. Similarly, the third baseline (denoted as ``vanilla+O") is reconstructed by adding the motion map into ``vanilla" and also fused these two features by modifying our MEA and MDA blocks. Table~\ref{tab:ablation_eff_dep_motion} summarizes the quantitative results of our network of three baselines. Apparently, ``vanilla+D-ED" and ``vanilla+O-ED" outperforms ``vanilla", which demonstrates that the depth map and the temporal motion map have their contributions to the success of our video RGB-D SOD network. Moreover, combining both the depth map and the temporal information together can further enhance the video RGB-D SOD performance of our network, as indicated by the superior MAE, F$_\beta$, and S$_\alpha$ scores of our ATF-Net over ``vanilla+D-ED" and ``vanilla+O-ED".

\vspace{2mm}
\noindent
\textbf{Cost paid for the performance improvements.}
In Table~\ref{tab:ablation_eff_dep_motion},
we further report the cost (i.e., the number of the model parameter, the training time, and the inference time) paid for the performance improvement by the depth map and the temporal information.
We can find that the number of model parameters increase from 24 M to 103 M, when we add the depth map or the temporal motion map into ``vanilla" (with the only RGB image). And the training time is increased from 6 hours to 9 hours due to the additional depth map or the temporal information, and the inference time is increased from 38 FPS to 12 FPS due  to the depth map or the temporal map. Furthermore, when our network utilizes both the depth map and the temporal map, the number of parameters, the training time and the inference time is increased from (103 M, 9 hours, 12 FPS) to (130 M, 11 hours, 10 FPS). Although we have paid some cost, our network also improves the video RGB-D SOD performance when we include the depth map and the temporal motion map.

\vspace{2mm}
\noindent
\textbf{Effectiveness of multiple inputs.}
Displayed in Table~\ref{tab:ablation_multi-inputs}, we have established five baseline networks to validate the efficacy of multiple inputs and fusion approaches. 
The initial baseline network (referred to as ``vanilla") is created by omitting the depth and optical flow branches from our ATF-Net architecture. This configuration is akin to a U-Net model and exclusively employs RGB images as inputs.
Next, we ascertain the most effective approach to integrate geometric information from depth images into the "vanilla" model by comparing various fusion methods. This involves straightforwardly concatenating depth images with RGB images along their channel dimensions (referred to as ``vanilla+D-concat"), or alternatively, incorporating an additional UNet-like branch (referred to as ``vanilla+D-branch").
Likewise, we formulate the ``vanilla+O-concat" and ``vanilla+O-branch" networks to determine the most optimal approach for leveraging temporal information from optical flow images.
As depicted in Table~\ref{tab:ablation_multi-inputs}, the incorporation of supplementary geometric and temporal information yields improvements in the efficacy of salient object detection in videos.

\section{Conclusions}
To facilitate the research on RGB-D video salient object detection (SOD),
we collected a new annotated RGB-D video SOD dataset (ViDSOD-100), which consists of 100 videos with 9,362 video frames covering various salient object categories.
Meanwhile, we devised an attentive triple-fusion network (ATF-Net) with four branches for RGB-D video SOD to integrate the appearance, temporal, and geometry information from the input RGB image, the input depth image, and an estimated motion map by embedding modality-specific modules.
Experimental results on ViDSOD-100 showed that our ATF-Net outperforms $21$ state-of-the-art methods in terms of RGB-D video SOD.
We hope our new dataset and benchmark would promote the development of RGB-D Video SOD community.
Further work includes the collection of more data in our ViDSOD-100. 

\section{Acknowledgements}
This work is supported by the Guangzhou Municipal Science and Technology Project (Grant No. 2023A03J0671), the InnoHK funding launched by Innovation and Technology Commission, Hong Kong SAR, the Guangzhou Industrial Information and Intelligent Key Laboratory Project (No. 2024A03J0628), the Guangzhou-HKUST(GZ) Joint Funding Program (No. 2024A03J0618), the Ministry of Science and Technology of the People’s Republic of China (STI2030-Major Projects2021ZD0201900), the National Natural Science Foundation of China (Grant No. 62371409), and National Research Foundation Singapore under its AI Singapore Programme (Award Number: AISG2-GC-2023-007).

\bibliography{ref}



\end{document}